\documentclass[lettersize,journal]{IEEEtran}
\usepackage{amsmath,amsfonts}
\usepackage{algorithmic}
\usepackage{algorithm}
\usepackage{array}
\usepackage[caption=false,font=normalsize,labelfont=sf,textfont=sf]{subfig}
\usepackage{textcomp}
\usepackage{stfloats}
\usepackage{url}
\usepackage{verbatim}
\usepackage{graphicx}
\usepackage{cite}
\usepackage{booktabs} 
\usepackage{pifont}
\usepackage[justification=centering]{caption}
\usepackage[numbers,sort&compress]{natbib}
\begin{document}

\title{Brain-inspired AI Agent: The Way Towards AGI}

\author{
  Bo Yu,
  Jiangning Wei,
  Minzhen Hu,
  Zejie Han,
  Tianjian Zou,
  Ye He,
  Jun Liu,~\IEEEmembership{Member,~IEEE,}
}

\maketitle

\begin{abstract}
Artificial General Intelligence (AGI), widely regarded as the fundamental goal of artificial intelligence, represents the realization of cognitive capabilities that enable the handling of general tasks with human-like proficiency. Researchers in brain-inspired AI seek inspiration from the operational mechanisms of the human brain, aiming to replicate its functional rules in intelligent models. Moreover, with the rapid development of large-scale models in recent years, the concept of agents has garnered increasing attention, with researchers widely recognizing it as a necessary pathway toward achieving AGI. In this article, we propose the concept of a brain-inspired AI agent and analyze how to extract relatively feasible and agent-compatible cortical region functionalities and their associated functional connectivity networks from the complex mechanisms of the human brain. Implementing these structures within an agent enables it to achieve basic cognitive intelligence akin to human capabilities. Finally, we explore the limitations and challenges for realizing brain-inspired agents and discuss their future development.
\end{abstract}

\begin{IEEEkeywords}
AGI, brain-inspired AI, agent.
\end{IEEEkeywords}

\section{Introduction}
\IEEEPARstart
{A}{GI} refers to creating (semi-)autonomous and adaptive computer systems with the general cognitive capabilities typical for humans, including the ability to support abstraction, analogy, planning and problem-solving \cite{ref1}. The concept of artificial intelligence dates back to the 1950s, when the term ``artificial intelligence'' was introduced during the 1956 Dartmouth Summer Research Project \cite{ref2}. Since then, researchers have persistently explored the feasibility of achieving AGI \cite{ref3,ref4,ref5}, despite numerous failed attempts. Over time, the increasing complexity of AGI led to a shift during the 1970s and 1980s from general-purpose intelligent systems to domain-specific problem-solving approaches. The introduction of the back-propagation algorithm \cite{ref6} in the 1980s provided the foundation for modern deep learning networks, such as Convolutional Neural Networks (CNNs) \cite{ref7} and Recurrent Neural Networks (RNNs) \cite{ref8}. The Transformer model, which leverages self-attention mechanisms, has become the foundation of a number of state-of-the-art artificial neural networks, such as BERT \cite{ref9} and GPT  \cite{ref10}. The emergence of these models signifies the explosive growth of modern artificial intelligence.  

To date, the artificial intelligence systems discussed above remain within the domain of special purpose AI. In fact, research of general-purpose artificial intelligence systems, considered the ultimate goal of AI, was revitalized at the turn of the 21st century and has become a central theme in scholarly discourse \cite{ref11,ref12,ref13}. The concept of AGI was further developed in 2007 by Ben Goertzel in his book \cite{ref14}, which has since garnered significant attention. Recent breakthroughs in deep learning, particularly with the advent of large language models, have reignited interest in AGI, with these models being regarded as sparks to AGI. Architectures based on large language models are widely considered a critical step toward the realization of AGI \cite{ref15,ref16}.  

The concept of an AI agent, introduced in the 1980s, refers to a framework within artificial intelligence designed to exhibit intelligent behavior, defined by characteristics such as autonomy, reactivity, pro-activeness, and social ability \cite{ref17}. With the rapid advancements in deep learning and the emergence of large language models \cite{ref18}, agents have increasingly been recognized as a pivotal area for the future trajectory of AI research. In recent years, the application of agents has expanded significantly across diverse fields, including autonomous driving \cite{ref19,ref20}, game AI \cite{ref21}, autonomous robotics \cite{ref22,ref23}, and intelligent healthcare \cite{ref24}.

However, current agent architectures are still largely task-specific, with their workflows constructed based on the task characteristics and heavily reliant on the processing capabilities of large language models (LLMs). There remains a significant gap between the capabilities of these architectures and the cognitive processing capabilities of the human brain. Although some agent architectures have incorporated elements that mimic brain structures to some extent, such as the Talker-Reasoner dual-system architecture proposed in \cite{ref25}, which simulates the brain's fast and slow thinking systems, or the integration of Theory of Mind (ToM) into Spiking Neural Networks (SNNs) as seen in \cite{ref26}, aimed at enhancing decision-making capabilities of multi-agent systems in both cooperative and competitive scenarios, notable limitations still persist.

While agent-based AI has made significant progress, brain-inspired AI has also advanced considerably. Recent studies have shown a strong similarity between Artificial Neural Networks (ANNs) and Biological Neural Networks (BNNs) \cite{ref27,ref28,ref29,ref30,ref31}, prompting many researchers to devote to this field. For example, \cite{ref32} was one of the first to propose using brain regions as model nodes with simple connection rules. The BI-AVAN network in \cite{ref33} mimics the human visual system's bias competition process to decode visual attention. Additionally, Spiking Neural Networks (SNNs) \cite{ref34} have been used to model neuronal behavior in the brain, effectively bridging the gap between brain-inspired AI and brain simulation \cite{ref35}. Despite significant advancements in both agent-based AI and brain-inspired AI fields, research has yet to focus on the design of agent architectures derived from brain-inspired structures.

Based on the aforementioned discussion, we propose that brain-inspired agent architecture offers a promising pathway toward realizing AGI. The perception-planning-decision-action (PPDA) model in agent architectures is typically interpreted in terms of human cognitive and behavioral patterns. Although significant progress has been made, this model and its extensions remain at a macro level of mimicking human cognition, without delving into the actual mechanisms of brain function. In fact, the human brain is an extraordinarily intricate structure, and its complex neural systems and brain region functional connectivity remain incompletely explored \cite{ref36,ref37}. Although we cannot directly model the neural circuits, we can shift our focus to the mesoscale cortical networks. In this architecture, we view different brain cortical areas as functional modules of the brain-inspired agent, with each module responsible for one or more tasks. Different functional modules cooperate to accomplish specific tasks, which researchers often associate with the small-world property of the brain \cite{ref38}. These functional modules, determined by task complexity and specific functional requirements, could be implemented using large language models (LLMs) or other appropriate tools. For instance, the V1 cortical area in the human brain is responsible for primary extraction of complex visual features, which can be implemented using CNN or YOLO \cite{ref39}; another example is the FPC region in the prefrontal cortex, one of the most complex areas in the human brain, responsible for higher-order cognitive functions, which we seek to emulate using LLMs. Additionally, the entire prefrontal cortex, as a large cortical area in the human brain, would be controlled by a core module and interact with other large cortical regions.   

Simultaneously, the agent can replicate the brain's process of extracting various types of information, with distinct cortical regions responsible for processing modality-specific features, such as visual, auditory, and olfactory, rather than being restricted to a single feature input. Upon entering the system, the information undergoes feature extraction in the relevant functional modules and is stored in the associated memory modules. Only when the brain region responsible for execution issues a command will the information be transmitted from the memory module to the designated brain region for analysis.

In the previous section, we discussed the functional implementation of brain-inspired agents. Another important aspect is how we reference the brain's cortical region connectivity mechanism, activating different regions and connection pathways for different types of tasks. According to \cite{ref40}, brain region connections are generally categorized into structural connectivity and functional connectivity. Our connectivity design primarily follows the functional connectivity model, which supports parallel task processing and cross-region information integration. In this connectivity model, the roles of different brain regions are clearly defined, the task execution sequence is explicit, and parallel task execution is possible. First, we simplify the complex functional connectivity of the human brain into regional connections between cortical areas. Each large cortical region contains multiple smaller cortical areas, with a core module responsible for their functional realization and interaction. The interactions between cortical regions facilitate the activation of distinct pathways corresponding to specific functional requirements. We argue that this brain-inspired connectivity mechanism has the potential to enable the agent to attain a level of intelligence that approximates human cognitive capabilities.

In summary, the widespread implementation of agent architectures marks a significant step forward in the pursuit of AGI. Inspired by the human brain, we propose a brain-inspired agent architecture designed to maximize the potential of agents and contribute to AGI. While the human brain, as the most intricate structure known, cannot be fully replicated in artificial intelligence, ongoing research and technological advancements, particularly the emergence of large language models, have renewed our hope of achieving general functionality in AI by progressively approximating the brain's architecture. The proposed agent architecture offers a promising framework for this endeavor, bringing us closer to the realization of AGI with cognitive abilities comparable to, or even surpassing, those of humans. Chapter \ref{sec:chapter2} will discuss research related to the human brain; Chapter \ref{sec:chapter3} will systematically analyze the development of agent architectures and their relationship to brain functions, validating the feasibility of our approach; Chapter  \ref{sec:chapter4} will address the limitations of the current method and propose directions for future research.

\section{Brain}
\label{sec:chapter2}
The brain, one of the most complex systems known, is responsible for regulating the body's physiological, cognitive, and behavioral functions. It processes sensory information, executes motor commands, supports higher cognitive functions like thought, decision-making, and memory, and controls the body's fundamental physiological processes. As the core of the central nervous system, the brain consists of multiple hierarchical structures. From a macroscopic perspective, it includes various components such as the cerebral hemispheres, cerebellum, and brainstem. On the microscopic level, the brain comprises over 86 billion neurons \cite{ref41}, each capable of forming up to 10,000 synapses with other neurons \cite{ref42}, resulting in a remarkably intricate network of connections that facilitates the emergence of intelligence. As a highly hierarchical structure, the brain is challenging to be modeled solely from either macroscopic or microscopic perspectives. In fact, at the mesoscale level of the human brain, different cortical regions are functionally independent but interconnected, with highly efficient functional connectivity mechanisms \cite{ref43,ref44}. However, owing to the complexity of the brain's structural organization, studying the distribution of cortical regions has been a key focus of biological research \cite{ref40,ref45,ref46}.

\subsection{Brain Cortex}
The cerebral cortex, the outermost structure of the brain, is divided into two hemispheres, each containing four primary functional areas: the frontal lobe, parietal lobe, occipital lobe, and temporal lobe. Each area is responsible for distinct functions and collaborates with others. The regions within each lobe can be further subdivided into smaller cortical areas, with a core module responsible for controlling the functions of the lobe and managing interactions with other lobes. Taking the frontal lobe as an example, the prefrontal cortex (PFC), as its core module, is responsible for decision-making, regulating planning, execution, and task management \cite{ref47}. The parietal lobe is responsible for processing sensory information (e.g., touch and spatial orientation) related to the body. The superior parietal lobule and inferior parietal lobule serve as its core control modules, with the prefrontal cortex (PFC) integrating this sensory information through interaction with them to guide complex motor behaviors or decisions \cite{ref48}. The interaction between the frontal and occipital lobes is related to visual perception and cognition, with visual input processed through the primary visual cortex, while the PFC helps formulate action decisions \cite{ref49}. The connection between the frontal and temporal lobes plays a vital role for memory processing, particularly for language, facial recognition, and the retrieval of long-term memories. In addition, the PFC itself can be divided into multiple core modules, including the Dorsolateral Prefrontal Cortex (DLPFC), Ventromedial Prefrontal Cortex (VMPFC), and Anterior Cingulate Cortex (ACC), with each module further subdivided into different functions \cite{ref50}, such as cognitive flexibility, risk assessment, and emotional regulation.

\subsection{Frameworks of Cerebral Cortex Division}
\subsubsection{Brodmann Areas Framework}
Brodmann Areas, proposed by German neuroanatomist Korbinian Brodmann in 1909, categorized different regions of the cerebral cortex based on structural differences, particularly the arrangement of cellular layers \cite{ref45}. Brodmann divided the cerebral cortex into 52 regions based on its microscopic anatomical features. Each area is assigned a number based on its morphological and functional characteristics, providing a framework for identifying specific brain regions. A key feature of the Brodmann Areas is that different regions typically perform different functions, although some of these functions may be revised as research progresses. Although the Brodmann Areas were not defined using modern neuroimaging techniques (such as fMRI), they provide an important theoretical framework for understanding the functional organization of the brain.
\subsubsection{Human Connectome Project Framework}
Human Connectome Project (HCP), proposed by Van Essen \textit{et al.} in 2012, used multimodal magnetic resonance images from the Human Connectome Project and combined an objective semi-automated neuroanatomical approach to delineate 180 regions per hemisphere in the precisely aligned group average of 210 healthy young adults. These regions are defined by sharp changes in cortical architecture, function, connectivity, and topography. The method characterizes 97 new regions and 83 previously reported regions identified using post-mortem microscopy or other specialized study-specific approaches, resulting in the final delineation of 180 regions. By integrating multimodal MRI techniques with a semi-automated neuroanatomical approach, HCP provides a new perspective on the parcellation of the brain cortex and offers important foundational data for understanding the brain's complexity and functional networks.
\subsubsection{Brain Cortical Region Connectivity Mechanism Framework}
The network connectivity mechanisms between cortical areas of the brain can be divided into structural connectivity and functional connectivity \cite{ref51}. Structural connectivity is established through anatomical neural fiber bundles (e.g., white matter tracts), which connect the brain regions via axons. These connections are static, remaining unchanged with fluctuations in brain activity. Major structural connections include local circuits, long-range fiber pathways, and cortico-cortical axonal pathways \cite{ref52}, which establish the fundamental framework for the brain's network operations. Functional connectivity refers to the activation of brain cortical regions under different functions, with these typical functional network pathways playing a crucial role in the brain's cognitive functions and being widely studied \cite{ref53,ref54,ref55}. The most well-known is the Default Mode Network (DMN), which is typically activated during rest, reflection, or memory processing. It primarily includes regions such as the medial prefrontal cortex, posterior cingulate cortex, hippocampus, and inferior temporal cortex. Another key network is the Executive Control Network (ECN), which is responsible for higher-order cognitive tasks such as working memory, decision-making, problem-solving, and planning. Other important functional networks include the emotion regulation network, auditory-visual network, motor network, language network, and self-related network.

Based on the aforementioned cortical division methods and functional connectivity mechanisms, we can selectively model the mesoscale operational mechanisms of the brain into the agent architecture. Different cortical regions of the brain are designed as functional modules based on their functional properties, each implementing specific functions. Upon receiving corresponding task instructions, key functional connectivity mechanisms and their associated functional modules will be activated. The goal is to construct a brain-inspired agent architecture and validate whether this structure can perform well on general cognitive tasks.

\section{Brain-inspired AI Agent}
\label{sec:chapter3}
\begin{figure}[!t]
  \centering
  \includegraphics[width=2.5in]{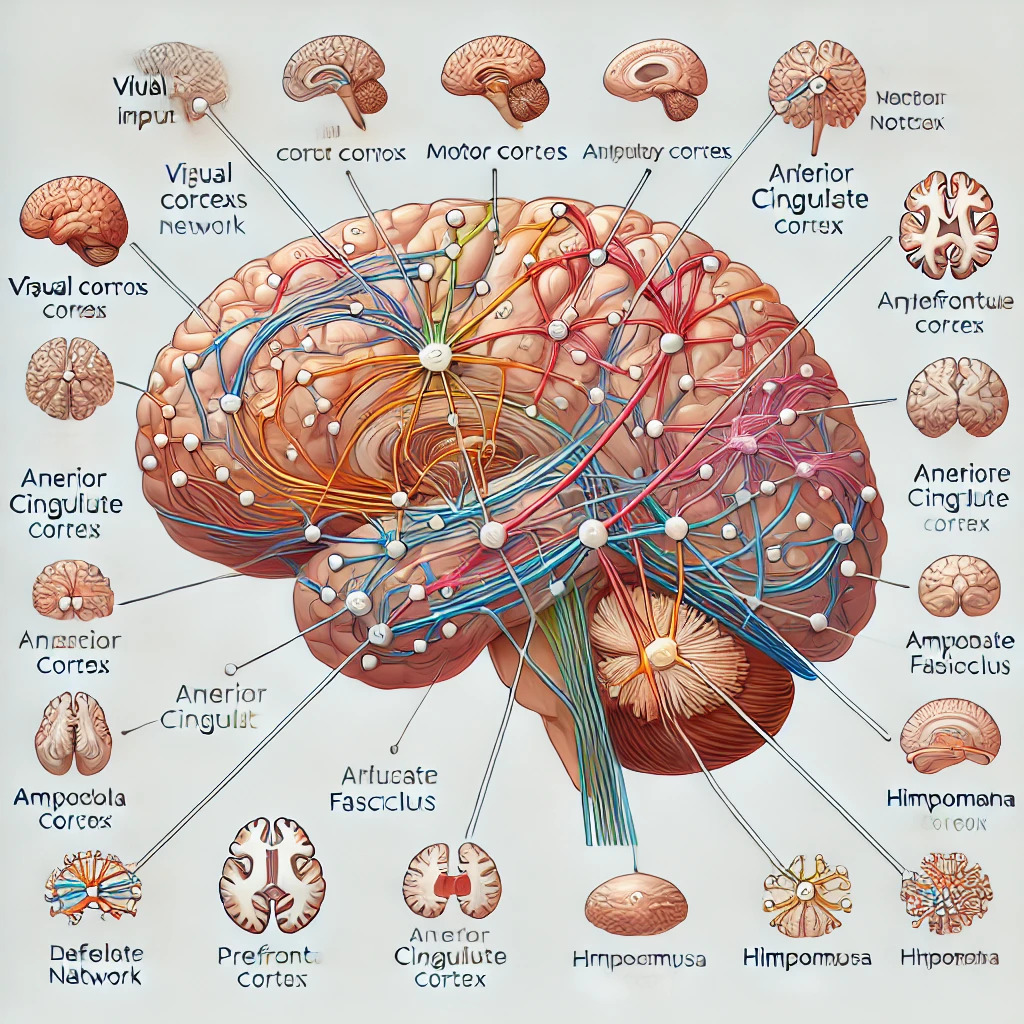}
  \caption{Schematic Diagram of Brain Regions in a Brain-Inspired Agent.}
  \label{fig_1}
  \end{figure}

Agent is considered a pivotal step towards AGI, and the rapid development of large language models in recent years has drawn significant attention to Agent. However, a unified agent architecture has not yet been well studied and established, and current agent structures and workflows are tailored to specific target tasks, resulting in a lack of flexibility and generality. Therefore, we aim to define a brain-inspired Agent structure based on the functional partitioning and connectivity networks of mesoscale cortical regions. This structure incorporates other fundamental brain functions and corresponding functional connectivity networks to the classic Perception-Planning-Action framework \cite{ref56}, as illustrated in Figure \ref{fig_1}. In Chapter \ref{sec:chapter3.1} , we will describe in detail how our brain-inspired structure is defined. Additionally, in Chapter \ref{sec:chapter3.2}, we will analyze existing Agent architectures and compare the differences between current Agent models and the brain-inspired structure.

\subsection{Brain-inspired AI Agent}
\label{sec:chapter3.1}
\begin{table*}[!t]
  \caption{Summary of Cortical Area Nodes and Functional Connectivity Networks for Brain-Like Functions.\label{tab:table1}}
  \label{tab:booktabs}
  \centering
  \resizebox{\textwidth}{!}{%
    \begin{tabular}{l c c c}
    \toprule
    Agent Functional Module & Corresponding Core Cortical Area & Main Function & Corresponding Cortical Network\\
    \midrule
    Perception & Primary Visual Cortex, Primary Auditory Cortex & Visual and Auditory Perception & Visual Input-Visual Cortex Network, Auditory Input-Auditory Cortex Network\\
    
    Planning & Prefrontal Cortex & High-level Cognitive Functions, Planning & Prefrontal Cortex-Motor Cortex Network, Prefrontal Cortex-Parietal Cortex Network\\

    Decision-making	& Dorsolateral Prefrontal Cortex	& Logical Reasoning, Decision Execution &	Anterior Cingulate Cortex-Dorsolateral Prefrontal Cortex-Parietal Cortex Network\\

    Action	& Primary Motor Cortex &	Motor Planning, Motor Execution	Motor & Cortex-Motor Network\\

    Memory &	Hippocampus, Prefrontal Cortex &	Memory Retrieval, Memory Transformation &	Hippocampus-Neocortex Pathway, Prefrontal Cortex-Hippocampus-Parietal Pathway\\

    Reasoning &	Dorsolateral Prefrontal Cortex, Ventromedial Prefrontal Cortex &	Logical Reasoning, Causal Analysis	&Prefrontal   Cortex-Parietal Pathway, Prefrontal Cortex-Hippocampus Pathway\\

    Reflection	& Medial Prefrontal Cortex	& Self-reflection, Emotion Regulation	& Default Mode Network Pathway\\

    Optimization	& Dorsolateral Prefrontal Cortex &	Cognitive 
    Control, Task Optimization	& Frontal Cortex-Lateral Prefrontal Pathway\\

    Emotion	& Amygdala, Anterior Cingulate Cortex &	Emotion Recognition, Emotion Regulation &	Amygdala-VMPFC-Anterior Cingulate Cortex Pathway, Amygdala-Hippocampus Pathway\\

    Language &	Broca’s Area, Wernicke’s Area &	Language Production, Language Comprehension &	Language Network (Arcuate Fasciculus) (Broca’s Area-Wernicke’s Area)\\
    \bottomrule
    \end{tabular}  
  }
  \end{table*}
The brain-inspired AI agent, as an agent model based on the structure of the brain's cortex, moves away from the traditional concept of using large language models (LLM) as the brain of the agent, treating the agent as a machine-like intelligence. Instead, it is conceptualized as a brain-inspired intelligent entity capable of managing a broad spectrum of general tasks. Its capabilities include, but are not limited to, perception, planning, decision-making, action, memory, reflection, optimization, emotion, language, and other common brain functions. These capabilities represent the human-like intelligence features that current agents aim to achieve. Based on the above functions, we identified the main cortical areas of the brain associated with these functions as functional nodes within the brain-inspired agent. Additionally, we simplified the corresponding functional connectivity networks, restricting interactions solely to the existing functional nodes. Each functional node is configured as one or more functional modules for its primary function and external interactions. For example, the primary visual cortex is set as a Visual Language Model (VLM) along with other object detection models to extract sufficient visual information for subsequent analysis. Table \ref{tab:table1} summarizes the corresponding cortical area nodes and functional connectivity networks for each brain-like function. Furthermore, we set an activation state for each functional node at every moment. This state depends on whether the node is currently involved in processing tasks within the relevant functional connectivity network. 
\subsection{ Survey of Current Agent Architectures}
\label{sec:chapter3.2}
\begin{table*}[!t]
  \caption{Focus on LLM-Based Single-Agent Architectures Aligned with Brain-Inspired Agents.\label{tab:table2}}
  \label{tab:booktabs}
  \label{tab:superbiblatex}
  \centering
  \resizebox{\textwidth}{!}{%
    \begin{tabular}{l c c c c c c c c c c}
    \toprule
    Agent & Perception & Planning & Decision-making & Action & Memory & Reasoning & Reflection & Optimization & Emotion & Language\\
    \midrule
    GEAP\textsuperscript{\cite{ref57}} &   & \ding{51} & \ding{51} & \ding{51} & \ding{51} &   &   &   &   &  \\
    WebShop\textsuperscript{\cite{ref58}} & \ding{51} & \ding{51} & \ding{51} & \ding{51} &  &   & \ding{51}  &   &   &  \\
    AT-agent\textsuperscript{\cite{ref59}} &   & \ding{51} &  &  & \ding{51} & \ding{51}  & \ding{51}  &   &   & \ding{51} \\
    LLift\textsuperscript{\cite{ref60}} &   & \ding{51} &  &  & &  &  &   &   & \ding{51} \\
    Intelligent-Agent\textsuperscript{\cite{ref61}} &   & \ding{51} &  &  &  &   &   &   &   & \ding{51} \\
    ChemCrow\textsuperscript{\cite{ref62}} &  \ding{51} & \ding{51} & \ding{51} & \ding{51} &  &   &   &   &   &  \\
    ChatMOF\textsuperscript{\cite{ref63}} & \ding{51}  & \ding{51} & \ding{51} &  &  &   &   &   &   &  \\
    DEPS\textsuperscript{\cite{ref64}} & \ding{51}  & \ding{51} &  & \ding{51} &  &   & \ding{51}  &   &   &  \\
    Voyager\textsuperscript{\cite{ref65}} &   & \ding{51} &  & \ding{51} &  &   & \ding{51}  &   &   &  \\
    DECKARD \textsuperscript{\cite{ref66}} & \ding{51} & \ding{51} & \ding{51} &  & \ding{51} &   &   &   &   & \ding{51} \\
    Plan4MC\textsuperscript{\cite{ref67}} & \ding{51} & \ding{51} & \ding{51} &  &  &   &   &   &   &  \\
    PET\textsuperscript{\cite{ref68}} & \ding{51}  & \ding{51} & \ding{51} & \ding{51} &  &   &   &   &   &  \\
    WebAgent \textsuperscript{\cite{ref69}} &   & \ding{51} &  & \ding{51} &  &   & \ding{51}  &   &   &  \\
    InterAct\textsuperscript{\cite{ref70}} &   & \ding{51} &  &  &  &   &   &   &   & \ding{51} \\
    WebGUM\textsuperscript{\cite{ref71}} &   & \ding{51} &  &  &  &   & \ding{51}  &   &   &  \\
    RCI-agent\textsuperscript{\cite{ref72}} & \ding{51} &  &  &  &  & \ding{51} &  &  &  &  \\
    Synapse\textsuperscript{\cite{ref73}} &  & \ding{51} & \ding{51} &  &  &  & \ding{51} &  &  &  \\
    Mind2Web\textsuperscript{\cite{ref74}} & \ding{51} & \ding{51} & \ding{51} & \ding{51} &  &  &  &  &  & \ding{51} \\
    Agent-Pro\textsuperscript{\cite{ref75}} &  & \ding{51} & \ding{51} &  &  &  &  &  &  & \ding{51} \\
    Self-Contrast\textsuperscript{\cite{ref76}} &  &  & \ding{51} & \ding{51} & \ding{51} &  & \ding{51} & \ding{51} &  &  \\
    UALA\textsuperscript{\cite{ref77}} &  &  &  &  &  &  & \ding{51} & \ding{51} &  &  \\
    OSWorld\textsuperscript{\cite{ref78}} & \ding{51} & \ding{51} & \ding{51} &  &  &  & \ding{51} & \ding{51} &  &  \\
    RAG-agent\textsuperscript{\cite{ref79}} & \ding{51} & \ding{51} &  & \ding{51} &  &  & \ding{51} &  &  & \ding{51} \\
    \midrule
    Brain-inspired-agent & \ding{51} & \ding{51} & \ding{51} & \ding{51} & \ding{51} &  \ding{51}& \ding{51} & \ding{51} & \ding{51} & \ding{51} \\
    \bottomrule
    \end{tabular}  
  }
  \end{table*}
We have summarized and reviewed some of the agent architectures emerging in the past two years and discussed their relationship with brain-inspired structures, including but not limited to embodied agents, virtual world scene agents, logical innovation agents, web indexing agents, and multimodal agents. Table \ref{tab:table2} primarily focuses on LLM-based single-agent architectures, which are of the same dimension as the brain-inspired agent we propose.  
As illustrated in Table \ref{tab:table2}, currently implemented agent architectures share similar brain-inspired structural ideas. Agents are equipped with different brain-like capability modules depending on specific tasks, indicating that researchers consistently acknowledge the necessity of brain-inspired capabilities to augment agent performance beyond that achievable by LLMs alone. This approach yields significant results in handling specialized tasks but remains insufficient for advancing general artificial intelligence. Therefore, it is crucial to explore and implement an agent architecture with general brain-like functions. Additionally, we abandon the traditional task-oriented workflow approach and instead design the agent's workflow based on the brain's functional connectivity mechanisms. We posit that these advancements will establish a robust foundation for achieving AGI.

\section{Discussion and Future Work}
\label{sec:chapter4}
In this paper, we designed a brain-inspired agent architecture and proposed the concept of a general-purpose artificial intelligence agent by emulating the structure of the human brain, while exploring its feasibility. However, significant limitations and challenges remain before realizing a brain-inspired agent capable of human-level intelligence, including:
\subsubsection{ Limited Understanding of the Human Brain} Despite notable progress in neuroscience and brain-inspired intelligence, achieving a comprehensive understanding of the human brain and replicating its intelligent functions within AI continues to be a formidable challenge for researchers.
\subsubsection{ Insufficient Definition of Brain-Inspired Architecture} Our research primarily focuses on mesoscale cortical regions and their associated connectivity networks, with limited attention to subcortical areas and more fine-grained neural connections. We believe that these regions, along with their corresponding functional connectivity networks, are also crucial for achieving brain-inspired intelligence.
\subsubsection{ Computational Resource Consumption}Brain-inspired agents will inevitably require significant computational resources. Therefore, how to allocate these resources efficiently poses an unavoidable challenge for our work.
\subsubsection{ Framework Integration} Integrating numerous functional nodes and connectivity networks poses another critical challenge. We aim to transcend existing integrated frameworks by developing a novel, specialized framework designed specifically for brain-inspired agents.

The future of brain-inspired agents is promising, with the potential for rapid growth, similar to the recent rise of large language models (LLMs). It is foreseeable that the era of agents is approaching, making it timely to focus on agents capable of closely approximating human-like intelligence. Such agents will inject new energy and limitless possibilities into numerous industries. The realization of brain-inspired agents will not only provide a new paradigm for achieving AGI, but also serve as a essential reference for addressing various general intelligence tasks.

Moreover, advancements in fundamental computer science algorithms and hardware capabilities will continue to significantly impact the development of artificial intelligence. Technologies such as Retrieval-Augmented Generation (RAG) \cite{ref79}, Reinforcement Learning \cite{ref80}, and Imitation Learning \cite{ref81} have also played an essential role in advancing AI.

Ultimately, we will aim to refine the relevant components of our work in the future to maximize the realization of brain-inspired agent structures and meaningfully contribute to the achievement of AGI.

\bibliographystyle{IEEEtranN}
\bibliography{IEEEabrv,ref}

\newpage

\vfill

\end{document}